Pierre-Frederic Villard*, Thomas M. Waite and Robert D. Howe

Cosserat Rods for Modeling Tendon-Driven Robotic Catheter Systems

## Abstract

Tendon-driven robotic catheters are capable of precise execution of minimally invasive cardiac procedures including ablations and imaging. These procedures require accurate mathematical models of not only the catheter and tendons but also their interactions with surrounding tissue and vasculature in order to control the robot path and interaction. This paper presents a mechanical model of a tendon-driven robotic catheter system based on Cosserat rods and integrated with a stable, implicit Euler scheme. We implement the Cosserat rod as a model for a simple catheter centerline and validate its physical accuracy against a large deformation analytical model and experimental data. The catheter model is then supplemented by adding a second Cosserat rod to model a single tendon, using penalty forces to define the constraints of the tendon-catheter system. All the model parameters are defined by the catheter properties established by the design. The combined model is validated against experimental data to confirm its physical accuracy. This model represents a new contribution to the field of robotic catheter modeling in which both the tendons and catheter are modeled by mechanical Cosserat rods and fully-validated against experimental data in the case of the single rod system.

Introduction

The field of cardiac treatments is rapidly expanding to provide safer and more efficient treatments for cardiac conditions as heart disease continues to be the leading cause of death in the developed world [1]. One particularly promising field for minimally invasive cardiac treatments is the use of catheters. Catheters can provide minimally invasive solutions to a variety of cardiac conditions. Not only do they have applications in a variety of intervention procedures including ablations [2, 3], but they also have many applications in imaging and diagnostics [4]. Maneuvering these catheters precisely is essential to effective treatments, yet manually it is a difficult task given the limited maneuverability and precision of manual catheters. Robotic catheters can provide greater precision and dexterity in performing such precise procedures [5], but there is a need for an accurate modeling system to allow for increasingly precise catheter-based interventions.

*Corresponding author: **Pierre-Frederic Villard,** Université de Lorraine, CNRS, Inria, LORIA, F-54000 Nancy, France, e-mail: pierrefrederic.villard@loria.fr
**Pierre-Frederic Villard, Thomas M. Waite, Robert D. Howe: Paulson** School of Engineering and Applied Sciences Harvard University, Cambridge, MA 02138, USA

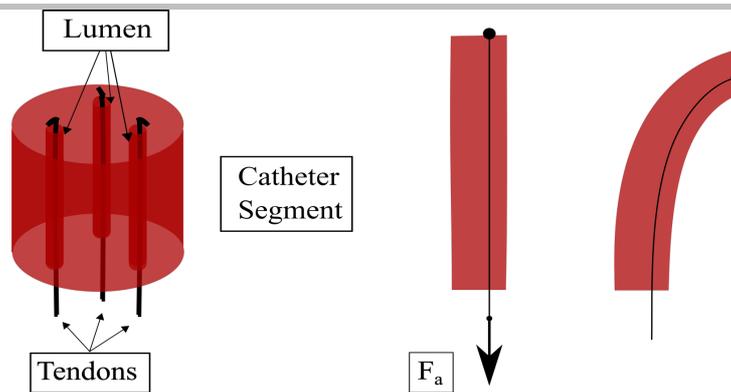

**Figure 1:** Tendon-driven robotic Catheter: (A) catheter segment composed of hollows cylinders (lumens) that constrain tendons, (B) shape of the catheter before and after applying an actuation force $F_a$ to one tendon.

Robotic catheters are composed of a set of actuation tendons restrained inside the lumens of a catheter body and attached to the catheter tip. When forces $F_a$ are applied to these tendons, the distal portion of the catheter bends (Fig. 1). The construction of these tendon-catheter system can include many tendons, thereby allowing many degrees of freedom in catheter motion. In addition to the mechanics of the catheter itself, it is also important that catheter models handle interactions with the patient [3,5,6]. Additional imaging techniques have been developed to aid surgeons in viewing catheters and tissues during the procedures [7], leading to precise data collection methods for robotic catheter systems [8]. These data collection systems can then be integrated into the control for the robotic catheter system, leading to even more precise guidance [3]. Robotic catheters also include the recent field of soft robotics with continuum manipulators made of soft materials that totally deform to better comply with the environment. This technology requires a continuum approach for the modeling and the simulation that could be done using the Cosserat formulation [9, 10]. However, such devices are not easy to manufacture.

1D deformable models have been applied to catheter simulation with various degrees of accuracy depending on the context. Most of the existing research deals with catheter insertion training simulators. In [11], the catheter is modeled with a mass-spring system, in which parameters are springs and damping coefficients. Catheters have also been modeled using the strict mechanical laws of beam theory [12]. A beam element is a straight bar of uniform cross-section with two extremities capable of resisting axial forces, shearing forces, and bending moments around the two principal axes in the plane of its cross-section as well as twisting moments or torques around its centroidal axis. In [13] the catheter and guidewire models consist of a series of nonlinear deformable beam elements. It includes collision detection and collision response. It is a simulator that only targets realism and therefore its accuracy has not been validated. This framework has been extended to include interactive fluid dynamics of blood flow [14]. Recent work has focused on computation efficiency and improving contact response with the blood vessel surface [15, 16]. However, this work does not include robotic assistance. It aims to be use for catheter manual insertion simulation. The mechanical properties are tuned to fit with a realistic behavior that is not precise enough in our context. In [17], the catheter is modeled with multi-section kinematics and is based on Bernoulli-Euler's hypothesis. The model has been validated with the use of real mechanical parameters. It does not include torsion.

Another mechanically-based solution to model 1D deformable rod is the Cosserat model. It has been used in animations as a means of modeling rods capable of self collisions. The main difference between Cosserat model and models from Euler-Bernoulli theory beam theory is that the effects of shear deformation could be simulated. One application is the animation of curly hair strands [18]. It has also been used in the medical context to model the spermatic cord for inguinal hernia repair simulation [19] or to model parallel continuum robot [20]. In the case of catheter modeling, Cosserat rods have been used extensively to model guide wires for vasculature and cardiac insertions [21, 22, 23, 24], for torque estimation in haptic devices [25] and for force and stiffness sensors in ablation applications [26]. In both of these applications an essential component is modeling collisions with vasculature or heart tissue. Cosserat rods are particularly well-suited to these applications as they are based on the continuum mechanical approach. That is, modeling collisions is simply a matter of calculating the appropriate collision force then applying it to the point masses in the Cosserat rod [27]. Cosserat rods have also been used as components in complex continuum robot models [28, 29, 30, 31]. In these applications, a single Cosserat rod is often used to model the backbone, taking advantage of variable material properties along the rod to aid in modeling the concentric continuum tubes. They have not been used, however, to model a system of control tendons combined with a catheter centerline. Tendon-driven catheter has been modeled with Cosserat rod theory in [32] for ablation purpose. The computation is not real time and has not been validated to check the whole catheter shape.

As a new application of Cosserat rods, we present a model which uses a system of Cosserat rods to model both the tendons and the catheter body of a robotic catheter system. These individual rods are then tied together by a series of constraints to resemble the real system. The result is a physically-determined system that can be adapted for a variety of catheter and tendon materials and scales. The only inputs to the system are the physical properties of the tendons and catheter and any applied forces, and the output of the system is the physical conformation of both the tendons and the catheter body. More importantly, by maintaining the mechanics of the system, the model provides a framework for capturing mechanical interactions in future robotic catheter models. For example, this model can be easily adapted to account for friction between the tendon and lumen wall, account for extensibility of tendons and catheters, or model the collision of tendon-driven catheters with vasculature and heart tissue. The aim is not to build a simulator for training but a tool that could be used by a clinician with high predictability so he/she can plan a treatment.

## Cosserat Rod Background

This section briefly reviews the Cosserat Rod mechanical model. We will first define the continuous version, then the discrete version, based on Spillmann et al.'s CORDE framework [33].

### Mathematical Representation of the Continuous Cosserat Rod

The Cosserat rod is modeled as the ordered set of center line points $\vec{r}(\sigma) = (r_x(\sigma), r_y(\sigma), r_z(\sigma))^T$ parameterized by $\sigma \in [0,1]$. Each point is assigned a set of right-handed orthonormal basis vectors $\vec{d}_1(\sigma)$, $\vec{d}_2(\sigma)$ and $\vec{d}_3(\sigma)$ called directors with $\vec{d}_3$ parallel to $\vec{r'}$, the spatial derivative of the centerline (Fig. 2), so that

$$\frac{\vec{r}'}{||\vec{r}'||} - \vec{d}_3 = \vec{0} \qquad (1)$$

This constraint between the directors and the centerline will be used to couple the forces on the material frame (director basis) to the forces that act on the centerline segments.

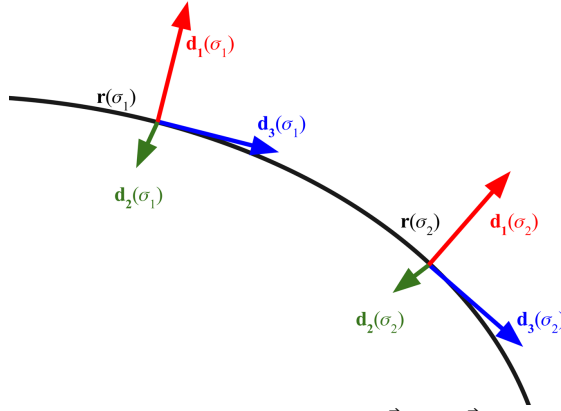

**Figure 2:** Cosserat centerline defined by $\vec{r}(\sigma)$ and director vectors $\vec{d}_1(\sigma)$, $\vec{d}_2(\sigma)$ and $\vec{d}_3(\sigma)$ represented at $\sigma = \sigma_1$ and $\sigma_2$

Rotation Definitions with Quaternions

A quaternion $\vec{q} = (q_1, q_2, q_3, q_4)^T$ represents the rotation of the rod. The directors $\vec{d}_k$ in terms of $\vec{q}$ are defined as

$$\vec{d}_1 = \begin{pmatrix} q_1^2 - q_2^2 - q_3^2 + q_4^2 \\ 2(q_1 q_2 + q_3 q_4) \\ 2(q_1 q_3 - q_2 q_4) \end{pmatrix}, \vec{d}_2 = \begin{pmatrix} 2(q_1 q_2 - q_3 q_4) \\ -q_1^2 + q_2^2 - q_3^2 + q_4^2 \\ 2(q_2 q_3 + q_1 q_4) \end{pmatrix}, \vec{d}_3 = \begin{pmatrix} 2(q_1 q_3 + q_2 q_4) \\ 2(q_2 q_3 - q_1 q_4) \\ -q_1^2 - q_2^2 + q_3^2 + q_4^2 \end{pmatrix} \qquad (2)$$

The quaternion has imaginary components $q_1, q_2, q_3$, real component $q_4$, and unity norm $||\vec{q}|| = 1$.

Now that we have defined the quaternions, we can rewrite the strain rates and angular velocities in terms of the quaternion. This is convenient for our model later since we can we be able to simply use the quaternion to calculate our energies directly.

$$u_k = \frac{2}{||\vec{q}||^2} \vec{B}_k \vec{q} \cdot \vec{q}' \qquad \omega_k = \frac{2}{||\vec{q}||^2} \vec{B}_k \vec{q} \cdot \dot{\vec{q}} \qquad \omega_k^0 = \frac{2}{||\vec{q}||^2} \vec{B}_k^0 \vec{q} \cdot \dot{\vec{q}} \qquad (3)$$

Here we also need to introduce the skew symmetric matrices $\vec{B}_k$. Their derivations can be found in Spillmann et al [31].

$$\vec{B}_1 = \begin{pmatrix} 0 & 0 & 0 & 1 \\ 0 & 0 & 1 & 0 \\ 0 & -1 & 0 & 0 \\ -1 & 0 & 0 & 0 \end{pmatrix}, \vec{B}_2 = \begin{pmatrix} 0 & 0 & -1 & 0 \\ 0 & 0 & 0 & 1 \\ 1 & 0 & 0 & 0 \\ 0 & -1 & 0 & 0 \end{pmatrix}, \vec{B}_3 = \begin{pmatrix} 0 & 1 & 0 & 0 \\ -1 & 0 & 0 & 0 \\ 0 & 0 & 0 & 1 \\ 0 & 0 & -1 & 0 \end{pmatrix} \qquad (4)$$

Similarly, we define the matrices with respect to the reference frame:

$$\vec{B}_1^0 = \begin{pmatrix} 0 & 0 & 0 & 1 \\ 0 & 0 & -1 & 0 \\ 0 & 1 & 0 & 0 \\ -1 & 0 & 0 & 0 \end{pmatrix}, \vec{B}_2^0 = \begin{pmatrix} 0 & 0 & 1 & 0 \\ 0 & 0 & 0 & 1 \\ -1 & 0 & 0 & 0 \\ 0 & -1 & 0 & 0 \end{pmatrix}, \vec{B}_3^0 = \begin{pmatrix} 0 & -1 & 0 & 0 \\ 1 & 0 & 0 & 0 \\ 0 & 0 & 0 & 1 \\ 0 & 0 & -1 & 0 \end{pmatrix} \qquad (5)$$

Energy Definitions

Strain rates are given by the spatial derivatives of the directors and angular velocities are given by the time derivatives of the vectors. They can be expressed in term of the quaternion to write the energies of the rod. With these energies, we will be able to calculate forces to evolve the rod system.
The potential energy of stretch is

$$V_s = \frac{1}{2}\int_0^1 K_s (||\vec{r'}|| - 1)^2 d\sigma \qquad (6)$$

where $K_s = E_s \pi r^2$ is the stiffness constant and $E_s$ is the Young's Modulus of stretching. In an ideal material, this is the same as the Young's modulus of bending $E_b$. Spillman et al. keep them distinct to introduce an extra degree of freedom, but they both correspond to the Young's modulus of the material to prevents free parameters in the material properties.
The potential energy of bending is

$$V_b = \frac{1}{2}\int_0^1 \sum_{k=1}^3 K_{kk} \left(\frac{2}{||\vec{q}||^2} \vec{B}_k \vec{q} \cdot \dot{\vec{q}} - \hat{u}_k\right)^2 d\sigma \qquad (7)$$

with stiffness tensor

$$K_{11} = K_{22} = E\frac{\pi r^4}{4},$$
$$K_{33} = G\frac{\pi r^4}{2}, \quad \text{and } K_{ij} = 0 \text{ otherwise} \qquad (8)$$

with G the shear modulus and r the radius of the rod's cross section.
The penalty energy is

$$E_p = \frac{1}{2}\int_0^1 K_p \left(\frac{\vec{r'}}{||\vec{r'}||} - \vec{d}_3\right) \cdot \left(\frac{\vec{r'}}{||\vec{r'}||} - \vec{d}_3\right) d\sigma \qquad (9)$$

This is a non-physical energy that maintains the constraint in Eq. (1), dictating that the quaternions are coupled to the centerline elements. Here, $K_p$ is a non-physical "spring constant" that dictates how tightly the quaternions are fixed to the centerline elements, which will be characterized below.

Discretization of the Rod

The centerline of $\vec{r}(\sigma)$ is discretized as a collection of N nodes. The centerline elements are thus defined as displacement vectors between control points, $\vec{r}_{i+1} - \vec{r}_i$. The orientations of each centerline element are defined by a quaternion attached to the midpoint of each of the rod elements (Fig. 3). Quaternions and centerline elements are linearly interpolated between adjacent elements. The energies for each segment are integrated over the length of each element. The derivative is computed with respect to each coordinate to determine the internal and external forces.

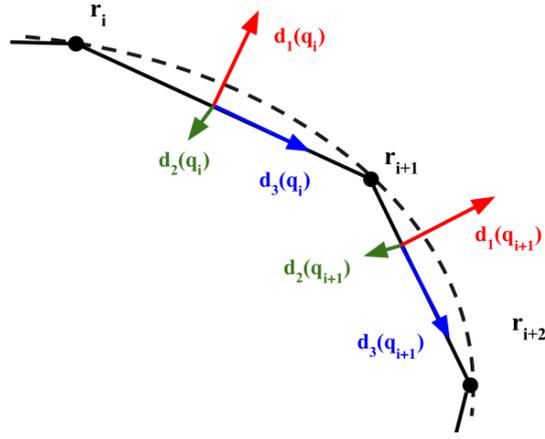

**Figure 3:** Discretized Cosserat rod: two segments are represented, $r_i r_{i+1}$ and $r_{i+1} r_{i+2}$, whose director vectors are respectively $\{\vec{d}_1(q_i), \vec{d}_2(q_i), \vec{d}_3(q_i)\}$ and $\{\vec{d}_1(q_i+1), \vec{d}_2(q_i+1), \vec{d}_3(q_i+1)\}$

Equations of Motion

The equations of motion for the control points are

$$\vec{a}_i = \dot{\vec{v}}_i = \frac{\vec{F}_i}{m_i} \quad (10)$$

$$\dot{\vec{r}}_i = \vec{v}_i \quad (11)$$

where $\vec{a}_i$ and $\vec{v}_i$ are the acceleration and velocity of the control point, $\vec{F}_i$ is the sum of internal and external forces, and $m_i$ is the mass of the control point. The force is calculated from the corresponding energies (Eq. (7) and (9)) by taking their partial derivative with respect to each of the coordinates in $\vec{r}_i$.

The equations of the motion for the quaternions are

$$\dot{\omega}_i = I^{-1}(\vec{\tau}_i - \omega_i \times I\omega_i) \quad (12)$$

$$\dot{\vec{q}}_i = \frac{1}{2} \vec{Q}_i \begin{pmatrix} 0 \\ \omega_i \end{pmatrix} \quad (13)$$

where $\tau_j$ is the sum of internal and external torques, $I$ is the inertia tensor, and $\vec{Q}$ is the quaternion matrix of q.

Integrating the Equations of Motion

Similarly to other work from the literature [11,13,14,15,19,24,33], we chose to solve the equations of motion within the dynamic case because it has better convergence properties and our goal is to navigate through a vascular network where a lot of contacts will occurs. For this model, an implicit integration scheme was chosen for simulation, which entails several advantages. Explicit integrators, while fast and easy to simulate, are inherently unstable when used to integrate stiff systems. The Cosserat model is particularly stiff both due to the stiff spring-like penalty forces used to enforce the parallel constraint from Eq. (1) and the spring-like energies dictating stretching and bending. These effects are especially severe when the scale of the system is small, as in the case of a catheter. For some applications, like animation, when speed is prioritized over physical accuracy and the scale is relative, an explicit integrator would be desirable. For present purposes, the stability and accuracy of an implicit system are important to model realistic catheters and tendons.

Using a semi-implicit Euler Method, we integrate the equations of motion (10) and (11), using time step h and a damping parameter $\xi \in [0, 1]$:

$$\vec{a}_i^t = \frac{\vec{F}^t}{m_i} \quad (14)$$
$$\vec{v}_i^{t+h} = \xi\vec{v}_i^t + h\vec{a}_i^t \quad (15)$$
$$\vec{r}_i^{t+h} = \vec{r}_i^t + h\vec{v}_i^{t+h} \quad (16)$$

to implement the Semi-implicit Euler integration scheme, we will first define the equations of motion for the system under an implicit integration scheme. The only difference from the explicit system is that now we use the position at the next timestep, $\vec{x}^{t+h}$, to determine the forces on the system. Thus, we can write the following definition of the Semi-implicit Euler equations of motion:

$$\vec{x}^t + \xi h \dot{\vec{x}}^t + h^2 M^{-1} \vec{F}_{net}(\vec{x}^{t+h}) = \vec{x}^{t+h} \quad (17)$$

where $h$ is the time step, $M$ is the mass matrix, and $\vec{F}_{net}$ is the force matrix. Here $\vec{x}^t$ is the column vector of coordinates at time $t$, where each of the 7 elements is the 3D coordinates and the quaternion for each point

$$\vec{x}^t = [x_1^t, y_1^t, z_1^t, q_1^t, q_2^t, q_3^t, q_4^t, \ldots, x_N^t, y_N^t, z_N^t, \ldots]^T \quad (18)$$

A damping parameter $\xi \in [0,1]$ has been added to allow variable damping of the system.
To apply this integration scheme, a solution to the system of equations must be computed. The $\vec{F}_{net}(\vec{x}^{t+h})$ term is nonlinear, so this is a root-finding problem, specifically the root of the function $\vec{f}$ where

$$\vec{f}(\vec{x}^{t+h}) = \vec{x}^t + \xi h \dot{\vec{x}}^t + h^2 M^{-1} \vec{F}_{net}(\vec{x}^{t+h}) - \vec{x}^{t+h} = 0 \quad (19)$$

To optimize the performance of the solver without fully defining the Jacobian matrix, we can simply define the sparsity array of the Jacobian. The chain-like structure of the Cosserat Rod means that the function of $\vec{f}$ at any coordinate only depends on those directly before and after, which results in the banded structure shown in Fig. 4. Mathematically,

$$\frac{\partial \vec{f}(\vec{x}^{t+h})_i}{\partial \vec{x}_j^{t+h}} = 0 \quad \forall j. \text{ s.t. } j \neq i, i+1, \text{ or } i-1 \quad (20)$$

This is a useful optimization as it results in $\approx 6 \times$ increase in solution speed for each timestep, given that hundreds or thousands of timesteps are often required for convergence.

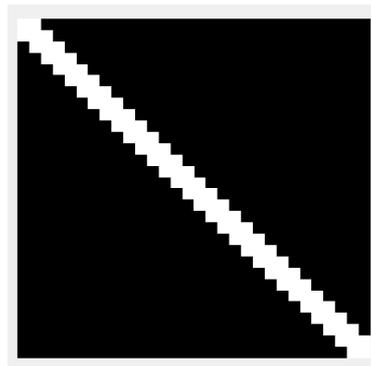

Figure 4: Banded sparsity structure of the single Cosserat model's jacobian matrix

## Single Rod System

This section will focus on the design and validation of a single Cosserat rod model against an analytical model for large deformation of cantilever beams and against experimental data from tests on a real catheter.

Analytical Model

Before considering the Cosserat model (CM) for a single rod subject to applied forces, we will implement an analytical model (AM) to compare for validation. This will allow us to validate the Cosserat Model for a wide variety of material properties, lengths, and geometries. The large deformation beam model of [34] describes the deflection of a cantilevered beam under an endpoint load $F$

$$\int_0^{\varphi_0} \frac{d\varphi}{\sqrt{\sin(\varphi_0)-\sin(\varphi)}} - 2\sqrt{\alpha} = 0 \quad (21)$$

$$x = \sqrt{\frac{2EI}{F}}\left(\sqrt{\sin(\varphi_0)} - \sqrt{\sin(\varphi)}\right) \quad (22)$$

$$y = \sqrt{\frac{EI}{2F}} \int_0^{\varphi} \frac{\sin(\varphi)d\varphi}{\sqrt{\sin(\varphi_0)-\sin(\varphi)}} \quad (23)$$

where $x$ and $y$ are the 2D coordinates of the the deformed body, $F$ is the load applied to the end of the rod, $I$ is the cross sectional area moment of inertia ($I = \frac{\pi r^4}{4}$ for a cylindrical beam), $E$ is the Young's Modulus, $\alpha$ is a dimensionless load parameter $\alpha = \frac{FL^2}{2EL}$, $L$ is the length of the rod, $\varphi_0$ is the resulting deflection angle of the end of the rod, and $\varphi$ is the deflection angle of any point along the rod. While having the advantage of expressing analytically the solution based on real parameters, this model has the drawback of ignoring shear deformation. The most likely differences are due to the other factors mentioned (intrinsic bending and gravity).

To solve these equations a large array of $\varphi_0$ values was defined and integrations were performed for each, solving for the corresponding $\alpha$ values. Then, we performed a simple lookup in the $\alpha$ array for our known $\alpha$ to find the $\varphi_0$ which generated it.

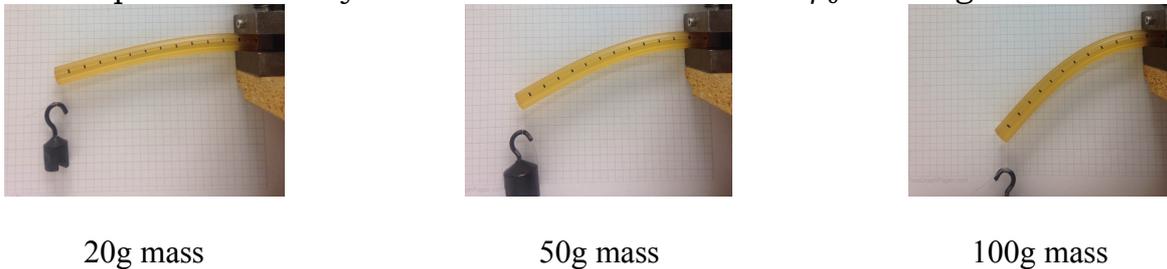

| 20g mass | 50g mass | 100g mass |

Figure 5: Experimental setup for cantilevered catheter. The catheter specimen has length 12 cm, radius 6 mm, and Young's Modulus 5.9 MPa. Three hanging masses are tested

Validation of Single Rod System

Experimental data for a catheter under a hanging mass load has been acquired to verify that the implementation of both the analytical model and the Cosserat model are physically accurate (Fig. 5). One end of the catheter was clamped and a hanging mass was affixed to the other end, providing a constant downward force.

The rod shape is planar and the idea is to take a picture it in a fronto-parallel plane. To make sure that it is in a front-parallel plane, a square was used as a test pattern to check that is was not deformed and no homography was needed to correct with the 4 corners. A graph paper was used the know size of the square to transcribe the pixel into millimeter positions. It was also necessary to ensure that the axes of the camera were well following the right vertical and horizontal, this was done by taking a plumb line.

The Cosserat and analytical models were computed in MATLAB (Mathworks, Natick, MA, USA). Accuracy was quantified as the difference in displacement of the endpoints between the reference analytical model and the Cosserat model normalized by the length of the rod, i.e. the percent error in tip displacement with respect to catheter length.

Fig. 6 shows a comparison of the experimental data, the Cosserat Model (CM), and the Analytical Model (AM). The data point corresponding to the experimental data are not completely smooth. This is due to human error as they were manualy extracted from the mm grid paper in the background. Fig. 7A and 7B show the influence of the penalty constant $K_p$ and the number of control points ($N$), respectively, on accuracy vs the analytical model. Fig. 8 shows the error against the analytical model over time for a number of damping values ($\xi$). In all of these situations, simulations of the Cosserat rod were run until convergence, that is, until the rod comes to rest after starting from rest in an undeformed state and evolving in response to a force which simulates a hanging mass on the endpoint. The full set of parameters for the Single Cosserat model that generated these results are displayed in Table 1.

**Table 1.** Single Catheter Experimental Parameters. The Young's modulus and the density are given by the catheter company, the radius and the length are measured, K_P, Δt, ξ and N are tuned.

| Property (Symbol) [Units] | Catheter |
|---|---|
| Young's Modulus of Bending ($E_b$) [MPa] | 5.9 |
| Young's Modulus of Stretching ($E_s$) [MPa] | 5.9 |
| Density ($\rho$) [kg/$m^3$] | 11040 |
| Radius ($r$) [m] | .006 |
| Length ($L$) [m] | .12 |
| Centerline Penalty Constant ($K_P$) | 1e4 |
| Timestep ($\Delta t$) [s] | .3 |
| Damping Constant ($\xi$) | .9 |
| Number of Control Points $N$ | 40 |

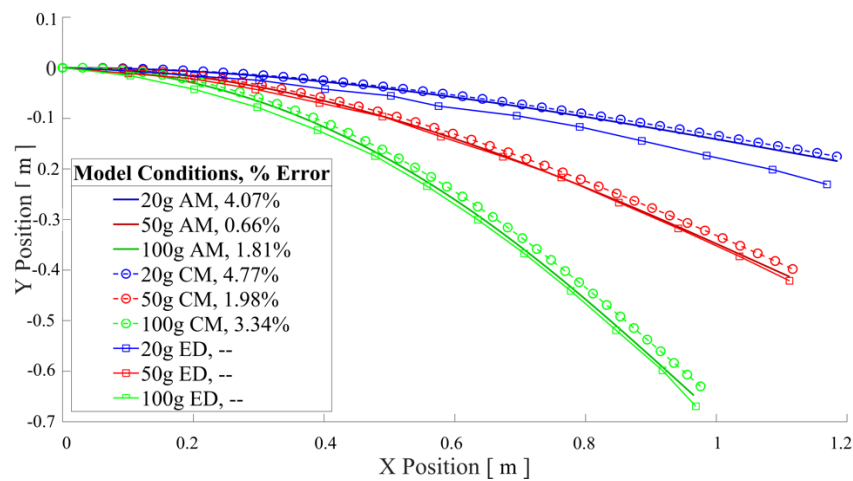

**Figure 6:** Comparison of Cosserat, Analytical, and Experimental Data. Control points are displayed at the equilibrium after applying the three hanging masses with the three methods

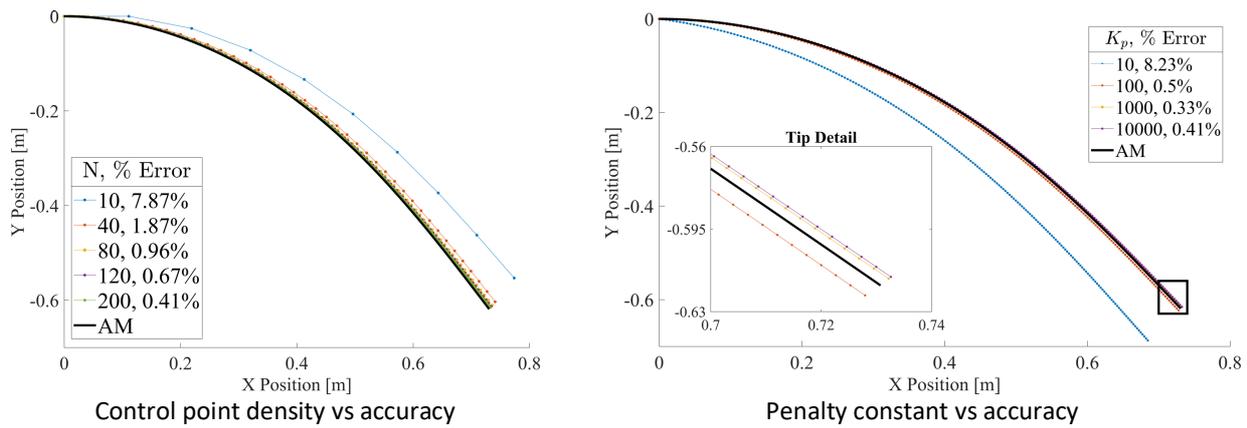

Control point density vs accuracy  Penalty constant vs accuracy

**Figure 7:** Analysis of the model accuracy on a 50N load

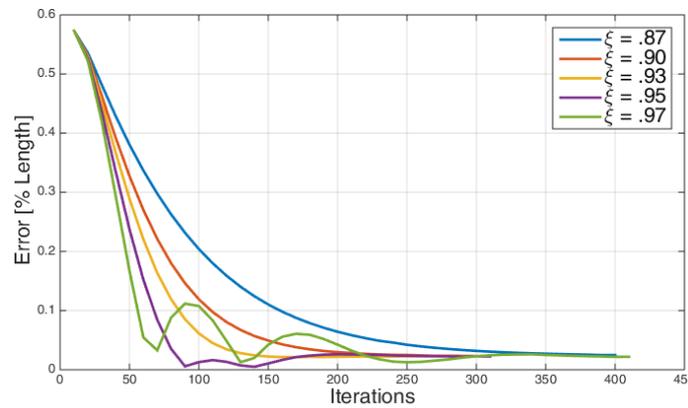

**Figure 8:** Error vs iteration number for various damping constant values

Discussion of Results

Overall, the Cosserat model fits well with both the experimental data and the analytical model (Fig. 6). While the Cosserat model is least accurate for small displacements against experimental data, we attribute this to some intrinsic bending in the catheter sample (violation of the cantilever boundary condition) as well as the influence of gravity. Regardless, the tip inaccuracies are all within 5% of the length of the rod, and in the moderate and large deformation case (20g and 50g applied), the error drops below 1% of the catheter length. The slight difference between the experimental data and the analytical model may come from the assumption of ignoring the shear deformability in the analytical model. All of the simulations in this section, including those on the experimental scale, run in ≈ 30 seconds on a standard desktop computer with 8 Gb of RAM. For simulations of very many control points or extremely high stiffnesses and penalty constants, that time would increase given the extra computation and the necessarily smaller timestep. However, we were able to achieve accurate results on all our physical scenarios without excessive computation times.

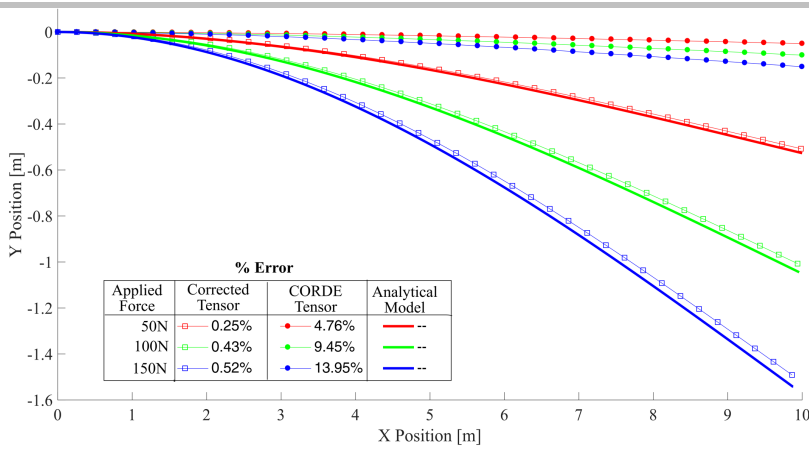

**Figure 9:** Comparison of corrected model and CORDE model against analytical model. Control points are displayed at the equilibrium after applying the three hanging masses with the three methods. The results with the original CORDE model fail to capture the large deformation.

Fig. 9 shows the results using both the stiffness tensor formula from [33] and our corrected value. From this test, we see that the CORDE version of the model is systematically overestimating the stiffness of the rods compared to the analytical model, leading to error in tip displacement of up to 90 % of the rod length in the largest deformation case. With this correction, the error between the Cosserat model and the analytical model dropped to under 3 % of the total length of the rod and remained nearly constant across applied forces.

As expected, the penalty constant which enforces the parallel constraint from Eq. (1) better satisfies the constraint with increasing value (Fig. 7A). Selecting a value that produces forces three orders of magnitude larger than the largest applied forces in the system is sufficient to ensure that the constraint forces are always significantly larger than the physical forces.

The purpose of the damping constant is to critically damp the system, so the model will converge as quickly as possible to its equilibrium position without oscillations. Several values are plotted with their error against the analytical model as function of iteration number (Fig. 8). A damping constant of $\xi = 0.90$ proved to be most effective for fast convergence and minimal oscillations. While this parameter may have additional implications for modeling the dynamics of the rod, it is only used here to minimize the simulation time.

Fig. 7B shows several Cosserat model rods with increasing control point counts ($N$) and their associated errors against the analytical model. increasing the number of control points effectively makes a better approximation of a real, continuous material. The errors are converging to 0, so as with the penalty constant, we simply need to pick a value large enough to provide sufficient accuracy. For these simulations, when $N \approx 40$, the error drops below 1% of the length of the rod. In conclusion, our implementation of the Cosserat model allows us to have a controlled accuracy depending on the control point density, the penalty constant value, the iteration number and the damping constant value. We will use this knowledge in the next section.

## Tendon-Catheter System

This section outlines the steps in defining the combined catheter-tendon system and design related to the constraint formulations. It also addresses issues and optimizations involved in simulating the expanded system. The results are then compared to experimental data. In order to simplify the formulation and testing, the tendons were initially located in a plane. The formulae that follow use this simplification, but the generalized formulation will be presented at the conclusion of this section and is supported by the design.

Implementation of Tendon-Catheter Constraints

The first step is defining constraints to tie together the tendon and catheter. We chose to implement these constraints with penalty forces analogously to how the centerline constraint was enforced in the CORDE system [33]. Applying the constraints is the same as altering our $F_{net}$ matrix of Eq. (19) on each iteration, which preserves much of the same simulation framework for the single rod.

Three distinct constraints must be formulated: the lumen compliance constraint, the endpoint compliance constraint, and the endpoint coupling constraint. Physically, the lumen compliance constraint corresponds to the requirement that the tendon remain inside the lumen of the catheter. The endpoint coupling constraint corresponds to the requirement that the tendon endpoint is a fixed distance from the catheter endpoint. And lastly, the endpoint compliance constraint will dictate how tightly the tendon endpoint is bound to the end of the lumen, as well as provide an extra free parameter to aid in fitting specific tendon-catheter systems. These constraints are illustrated in Fig. 10, which shows an example violation of the lumen constraint as well as the locations and directions in which endpoint constraints will act. The green region represents the lumen channel. Each of the constraints are highlighted in red. The larger black segments are the catheter centerline, and the smaller black segments represent the tendon centerline.

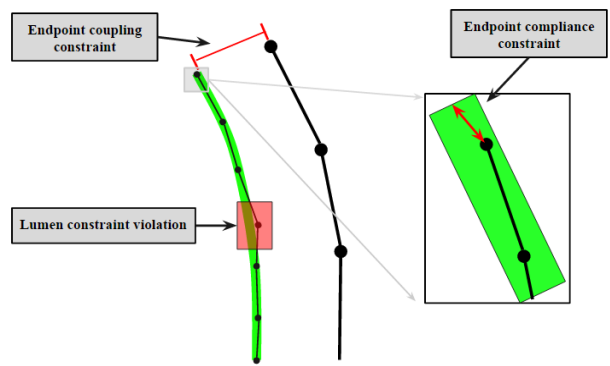

**Figure 10:** Tendon Catheter System Constraint Illustration

Lumen Constraint. The overarching goal of this constraint formulation is to define a vector of penalty forces that keep the tendon inside the lumen of the catheter. For this, we need to first make a representation of the lumen, the channel located a fixed distance from the centerline of the catheter body. Once we have a clearly defined lumen, we will need to define an appropriate penalty force to keep tendon points on the newly defined lumen centerline. In addition, the direction of these penalty forces will depend on where the tendon points lie along the length of the lumen. Thus, we will also need to register all the tendon points to a parent lumen segment then define a force that pushes them onto this parent lumen element.
To make an explicit definition for the lumen points, we can take advantage of the predefined orientation basis for each element of the Cosserat rod. We then follow the process outlined in Fig. 11: scaling the $\vec{d_1}$ vectors of each centerline element by the distance to the lumen from the centerline, $r_L$, then generating the lumen point by adding these scaled $\vec{d_1}$ to each of the catheter centerline points. For each point $\vec{r}_i$ on the catheter, then, we then generate each point on the lumen $\vec{L}_i$ as

$$\vec{L}_i = r_L \vec{d_1}_i + \vec{r}_i \qquad (24)$$

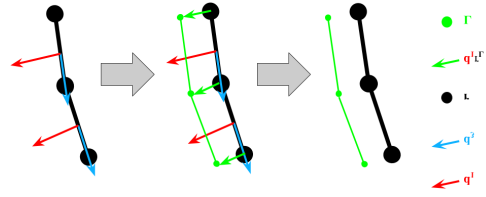

**Figure 11:** Lumen Generation Process

Next we must register each of the tendon points to exactly one parent lumen element. By registering to a parent, we will be able to apply the penalty forces on these tendon points in the direction of the parent catheter's $\vec{d_1}$ vector that generated the parent lumen element. This is illustrated in Fig. 12A, whereby the red and orange regions indicate approximately to which parent element each tendon point will be registered. In practice, this is implemented by finding the minimum distances from each tendon point to nearby lumen points, then recording the single parent element to which each tendon point belongs. This process ensures each tendon point is registered to exactly one parent element, and it needs to be recomputed on each iteration of the simulation because the location of both the tendon points and the lumen will change throughout the simulation. Thus tendon points can slide between parent elements regularly, especially when the control point density of the tendon is much larger than that of the catheter (Fig. 12).

After the tendon points are registered, the penalty force will be applied in the $\pm\vec{d_1}$ vector direction for a general point inside the lumen. The magnitude and precise direction of this force, however, will be dictated by a lumen compliance term $C_L$ which is a measure of how far the tendon points are outside of the lumen. The general form of this force $(\vec{F_L})_i$ will thus be

$$(\vec{F_L})_i = [K_L(C_L)_j](\vec{d_1})_j \qquad (25)$$

where $K_L$ is a penalty constant which will allow us to vary the severity of the constraint.

We next must define an appropriate metric for the compliance term. The important conditions for this metric are that it is 0 when the tendon points are satisfying the constraint (are on the correct lumen element) and that it is signed indicating the side of the lumen it lies on. For this, we will use the dot product as it is proportional to the signed scalar projection of the tendon points onto the lumen centerline. Specifically, we calculate displacement vectors between each of the tendon points and their nearest lumen points, and then calculate the dot product between these displacement vectors and the $\vec{d_1}$ vectors of the parent lumen element. Expressing this compliance term formally, let $\vec{p}_i$ be the position of tendon point $i$, and $\vec{L}_j$ be the nearest lumen point. This leaves the compliance term

$$(C_L)_j = (\vec{L}_j - \vec{p}_i) \cdot (\vec{d_1})_j \quad (26)$$

Substituting the compliance term into the force definition, we have:

$$(\vec{F_L})_i = [K_L(\vec{L}_j - \vec{p}_i) \cdot (\vec{d_1})_j](\vec{d_1})_j \quad (27)$$

The application of this process to many tendon points can be seen in Fig. 12B.

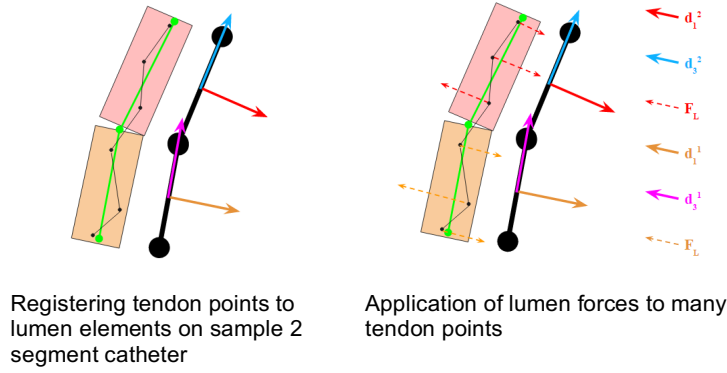

Registering tendon points to lumen elements on sample 2 segment catheter

Application of lumen forces to many tendon points

**Figure 12:** Tendon Point Registration and Penalty Force Application Process

With the lumen penalty forces determined, we need to incorporate these forces into the equations of motion for both the catheter and the tendon according to Newton's 1st and 3rd laws. One difficulty in applying those forces to the catheter points in particular is the different and variable numbers of tendon control points for each parent catheter centerline segment. However, because all tendon points are registered to a single parent catheter segment on each iteration, we are able to simply average the lumen forces on the tendon points over each parent centerline element on the catheter and apply the net force to the corresponding parent catheter control point.

While the forces above apply to all of tendon points inside the lumen of the catheter, we will define a distinct penalty force on the tendon endpoint. Defining this constraint separately is needed for two reasons. First, it introduces a free parameter which will allow us more flexibility in fitting individual tendon catheter systems. Secondly, without this constraint, we introduce an issue of interfering penalty forces, problem we will discuss in depth later.

Since we are now only dealing with the endpoint, we can simply define penalty force which tethers the tendon endpoint to the lumen endpoint. If $N_c$ is the number of control points in the catheter and $N_t$ is the number of points in the tendon, $\vec{t}_{N_t}$ is the endpoint of the tether and $\vec{L}_{N_c}$ is the endpoint of the lumen, the endpoint compliance penalty force is defined as

$$\begin{aligned}(\vec{F_E})_{N_t} &= K_E C_E \frac{(\vec{L}_{N_c} - \vec{t}_{N_t})}{||\vec{L}_{N_c} - \vec{t}_{N_t}||} \\ &= K_E ||\vec{L}_{N_c} - \vec{t}_{N_t}|| \frac{(\vec{L}_{N_c} - \vec{t}_{N_t})}{||\vec{L}_{N_c} - \vec{t}_{N_t}||}\end{aligned} \quad (28)$$

The force vector for this constraint can be seen in Fig. 13 as a part of the larger system, where the force vector and compliance factor are illustrated in blue.

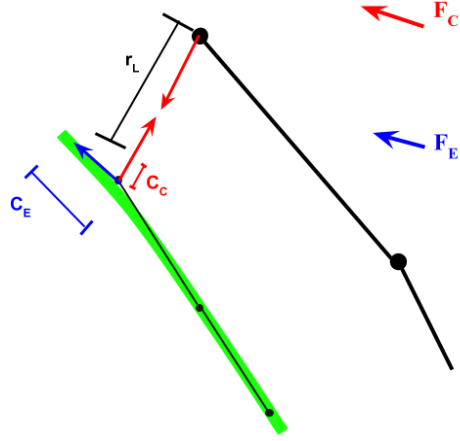

**Figure 13:** Diagram of Endpoint Penalty Forces

Endpoint Coupling Constraint. To tie the catheter centerline to the tendon centerline, we again implement a penalty force that enforces a displacement. However, instead of enforcing 0 displacement as we did with the endpoint on the tendon in the lumen, we will now enforce a resting displacement of $r_L$, the distance from the catheter centerline to the lumen. We define a compliance factor $C_C$ as the deviation from the resting distance, a new penalty constant $K_C$, and a direction vector, the normalized displacement vector between the tendon and catheter endpoints

$$\begin{aligned}\vec{F_c} &= K_C C_C \frac{(\vec{t}_{N_t} - \vec{r}_{N_C})}{||\vec{t}_{N_t} - \vec{r}_{N_C}||} \\ &= K_C (||\vec{t}_{N_t} - \vec{r}_{N_C}|| - r_L) \frac{(\vec{t}_{N_t} - \vec{r}_{N_C})}{||\vec{t}_{N_t} - \vec{r}_{N_C}||}\end{aligned} \qquad (29)$$

This force is applied to both the tendon and the catheter endpoint in opposite directions. The force vectors and the compliance term are illustrated in red in Fig.13.

Interference between Penalty Forces at Endpoints. Now that we have a fully constrained system, attention must be paid to the non-physical nature of the penalty forces we implemented. By their nature, some small displacement is required to maintain the constraint. In general, by setting the penalty constraints to be very large, we can minimize this displacement. However, because the endpoint of the tendon is doubly constrained by the endpoint compliance and coupling constraints, any displacement from the lumen endpoint will influence the amount of force applied for the coupling constraint and vice versa. To ameliorate this, one constraint is treated as fixed, then the other is modulated as a free parameter to fit the data. We fix the coupling constant and then use the compliance constant to do the fitting; while this could be done the other way, it is convenient to vary the compliance constant because it only acts on one control point, leading to more stable simulations. If we instead fixed the compliance constant and then had to set the coupling constraint to be very hard to fit the data, we are more likely to encounter the stiff spring problem outlined in section 2.6. We will sweep over values for both of these parameters below, however, and demonstrate their similar influence on the overall behavior of the system.

Integrating the Catheter-Tendon System.

After presenting the new constraints, the new tendon-catheter simulation framework is detailed in this section. It is based on the previous study of the single catheter system, simply expanding the system of equations to include both the tendon and the catheter centerline.

Formulation of Two-Body System. The new coordinates can be written as a vector, concatenating the coordinate vector from the tendon ($\vec{x_t}$) to the end of the vector for the catheter ($\vec{x_c}$)

$$\vec{x} = [\vec{x_c}^T \vec{x_t}^T]^T \qquad (30)$$

Similarly, the force and velocity vectors are concatenated, leading to the same formulation of a nonlinear system as in the single catheter case

$$\vec{f}(\vec{x}^{t+h}) = \vec{x}^t + h\xi\dot{\vec{x}}^t + h^2 M^{-1} \vec{F}_{net}(\vec{x}^{t+h}) - \vec{x}^{t+h} = 0 \qquad (31)$$

We note that the number of coordinates and the number of stiff spring-like forces increased dramatically from the single body case. There is thus a critical need for optimization, which will be addressed in the next section.

Defining the Jacobian Pattern. With a larger and stiffer system of equations to solve in the two-body case, it is vital to optimize the integration step by giving information on the Jacobian's sparsity pattern to the solver. There are new interdependencies between the coordinates now as a result of the lumen forces on the tendon depending on the poses of the catheter centerline points. Information on which coordinates were being used to derive the lumen force was saved, however, in the calculation of the lumen force. Therefore, we can readily define the new sparsity structures (Fig. 4A, B). This results in the same banded pattern as from the single rod along the main diagonal in both sparsity structures. This will remain constant throughout simulation as those are the relationships that determine the internal forces between coordinates such as stretching, bending, and the centerline penalty. The lower left band is accounting for the tendon's dependence on the coordinates of the catheter in the calculation of the lumen constraint forces, and the small square at the center right is accounting for the interdependence at the endpoints for the coupling constraint. As the relative control point densities or positions between the catheter and the tendon change, so too does the sparsity structure as tendon points move to new "parent" elements on the catheter. Thus, this sparsity structure needs to be redefined on every iteration, but again, the information needed to generate it is readily available from the force calculations and is thus an inexpensive computation.

Results for the Tendon-Catheter System

This section will first characterize the influence of the new parameters on the tendon catheter system. Then, using this information, we will fit and compare the tendon catheter model to experimental data collected from a precise robotic catheter visual capture system [8].

Analysis of Control Point Densities. Similar to the single rod case, as the number of control points in catheter centerline increases, the result will become more physically accurate as it better approximates the continuous material of the real catheter. Given that the interactions between the tendon and the catheter are dictated entirely by the endpoints, so long as the tendon points are constrained to the lumen, their relative density should have little impact on the overall behavior of the system. Of course, if we wanted to determine precise information about the tendon's deformation in the lumen or the exact extension of the tendon out of the lumen as a result of applied force, we would need a similarly high density to the catheter for physical accuracy. To test these behaviors, we plot the centerline of the catheter after equilibrium for a simulation in which a constant load is applied to the tendon. Fig. 14 shows the result of running this simulation with a 1:1 ratio of catheter:tendon control points.

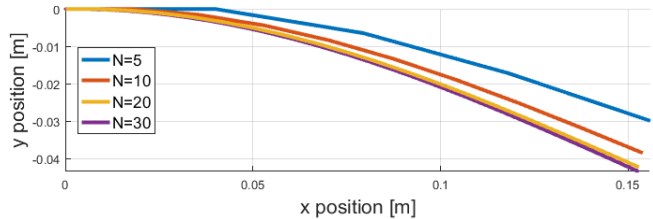

**Figure 14:** Comparison of Control Point Densities in Catheter-Tendon System

As the number of control points increases, the catheter tendon system rapidly converges. The next test varied tendon control points densities with the catheter control point density fixed at $N_C = 30$, the value we expect to be sufficient for physical accuracy (Fig. 15). As expected, the control point density of the tendon has practically no influence on the deformation of the rod. More points on the tendon may over-constraint the problem but it is already constrained by the high penalty force. For the remainder of the tests, then, we will fix $N_T = 10$.

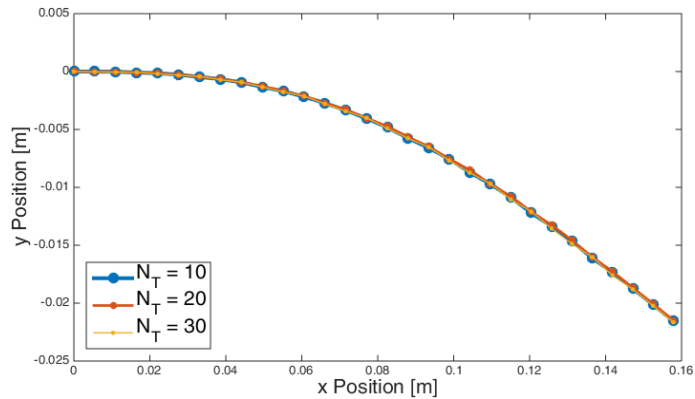

**Figure 15:** Comparison of Tendon Control Point Density for Fixed Catheter Control Point Density

**Analysis of Lumen Penalty Constant.** Similarly to the penalty constant in the single rod system $K_P$, we expect the lumen penalty constant to have little influence on the overall system above a certain threshold. This is confirmed by simulating a rod with $N_C = 30$, $N_T = 10$ and various $K_L$ values in Fig. 16. Above a penalty constant value of 1000, the lumen constraint is effectively satisfied, and further increase to 1300 does little to change the overall deformation for the system, so $K_L = 1000$ is used for the remainder of the tests. As before, picking the smallest penalty constant that effectively satisfies the constraint avoids unnecessary stiffness in the system.

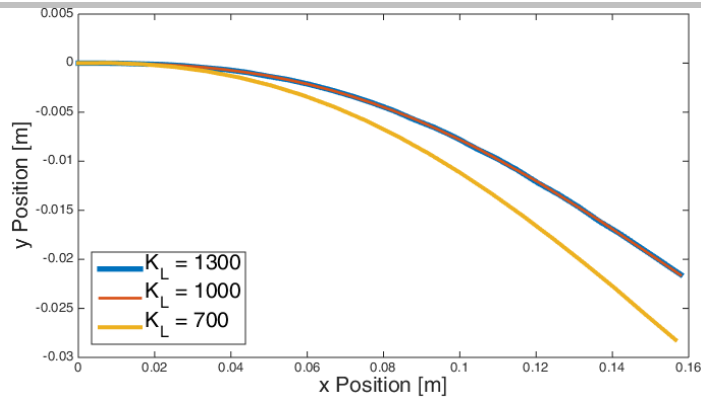

**Figure 16:** Comparison of Lumen Penalty Constant Values

Analysis of Endpoint Coupling and Compliance Penalty Constants. Next we will analyze the influence of the coupling penalty constant. As stated previously, we will eventually fix this parameter and use the endpoint compliance constraint for fitting, but we will here analyze both of their individual influences to prove that they have similar effects on the overall deformation and thus can be used interchangeably as free parameters. In Fig. 17, we sweep across $K_C$ values to determine the influence of the constant on the overall deformation. For this parameter sweep, we take the number of control points to be $N_C = 30$ and $N_T = 10$ and we fix $K_E = 1500$. The general behavior of this penalty constant is to control the amount of effective force applied to the catheter. As noted before, there needs to be some displacement for the penalty forces to take effect; thus, by increasing the value of the coupling constraint, we are essentially applying more force to the catheter endpoint for the same applied force to the tendon.

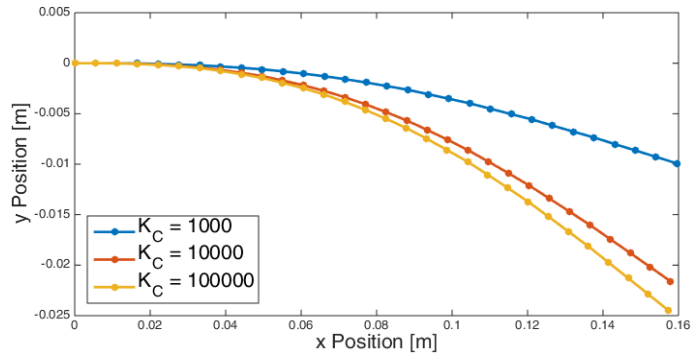

**Figure 17:** Comparison of Endpoint Coupling Penalty Constant

A similar sweep across $K_E$ values maintaining the same control point densities and holding $K_C = 100000$ are shown in Fig. 18. Similarly, the endpoint compliance term influences the amount of force translated to the endpoint of the catheter. The relationship is different than that of the coupling constant, however, in that as the value of $K_E$ decreases (the lumen becomes more compliant), more force is translated through the coupling forces to the catheter because the displacement between the endpoint increases. A second difference from the $K_C$ case is that the displacements that results from changing $K_E$ are almost linear in the value of the constant. In contrast, Fig. 17 indicates an more convergent behavior with increasing $K_C$ values. This is yet another reason that $K_E$ makes a better free parameter; its influence is predictable.

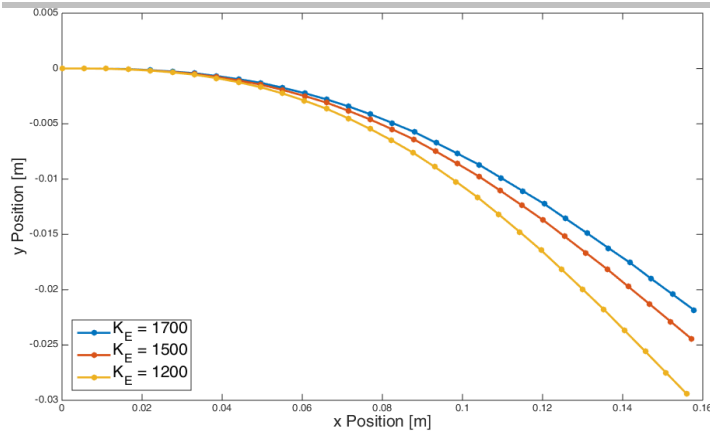

**Figure 18:** Comparison of Endpoint Compliance Penalty Constant

**Table 2.** Experimental Catheter Properties

| Property (Symbol) [Units] | Catheter |
|---|---|
| Young's Modulus of Bending ($E_b$) [MPa] | 5.9 |
| Young's Modulus of Stretching ($E_s$) [MPa] | 5.9 |
| Density ($\rho$) [kg/$m^3$] | 11040 |
| Radius ($r$) [m] | .006 |

Comparison to Experimental Data. With the control point densities and the new penalty constants characterized, we can now compare our model to the result from experimental data. The data is from a precise, visually-captured robotic catheter setup with a single tendon, as described in detail in [8]. The material properties of the catheter and tendon used are summarized in table 2. The only difference in the catheter properties between the tendon-catheter experiments and the hanging mass experiment is its length. In the hanging mass experiments, it was 12 cm. In the tendon experiments it was 16 cm. The experimental setup is shown in Fig. 19. The experimental condition of the data we use to compare was a force of $F_A = 2N$ applied to the tendon.

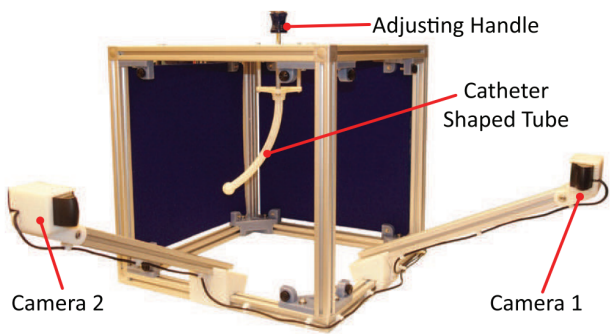

**Figure 19:** Experimental Setup for Tendon Catheter System of [8]

The application of the lumen forces to the catheter centerline to a sample simulated catheter and tendon is shown in Fig. 20. Without these forces, the comparison to the experimental data revealed a systematic error related to predicting the proper curvature of the catheter. The application of these forces to a sample simulated catheter and tendon is shown in Fig. 20.

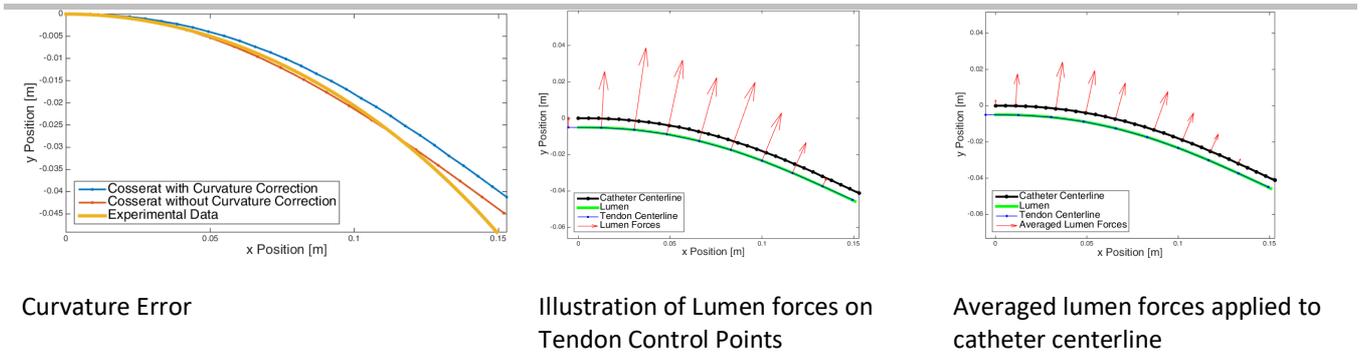

| Curvature Error | Illustration of Lumen forces on Tendon Control Points | Averaged lumen forces applied to catheter centerline |

**Figure 20:** Application of averaged lumen forces to catheter centerline

Because the endpoint error measure cannot differentiate accuracy based on the curvature of the rods, we define a more comprehensive error measure that can account for curvature problems, specifically the area between the catheter centerline curves normalized by the length of the rod. An example of the calculation of this error is show in Fig. 21, implemented by calculating the area of the polygon formed by the model and experimental data, then dividing by the length of the rod. One should note that even if the methods are in 3D, the area error is only computed in 2D as a simplification because we are in a fronto-parallel plane.

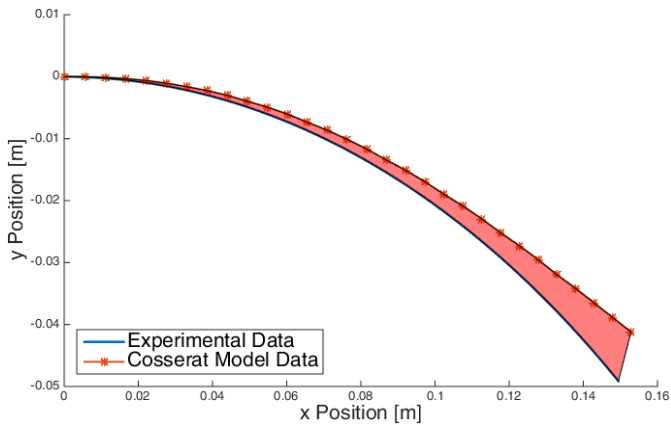

**Figure 21:** Improved Error Calculation

**Table 3.** Catheter-Tendon Fitting Parameters

| Property (Symbol) [Units] | Catheter 1 | Catheter 2 | Catheter 3 | Tendon |
|---|---|---|---|---|
| Young's Modulus of Bending ($E_b$) [MPa] | 5.9 | 5.9 | 5.9 | 5.9 |
| Young's Modulus of Stretching ($E_s$) [MPa] | 5.9 | 5.9 | 5.9 | .001 |
| Density ($\rho$) [kg/$m^3$] | 11040 | 11040 | 11040 | 10000 |
| Radius ($r$) [m] | .006 | .006 | .006 | .0001 |
| Length ($L$) [m] | .16 | .16 | .16 | .16 |
| Endpoint Compliance Constant ($K_E$) | 950 | 1100 | 1200 | N/A |
| Endpoint Coupling Constant ($K_C$) | 2e5 | 1e5 | 1e5 | N/A |
| Lumen Constant ($K_L$) | 1000 | 1000 | 1000 | N/A |
| Timestep ($\Delta t$) [s] | .2 | .2 | .2 | .2 |
| Damping Constant ($\xi$) | .9 | .9 | .9 | .9 |
| Number of Control Points $N$ | 30 | 30 | 30 | 10 |

We now assess the accuracy of our model against the experimental data under a variety of fitting conditions. Fig. 22A shows a plot of the catheter centerlines against the experimental data, and Fig. 22B shows the corresponding errors. Under the proper fitting conditions, our normalized error measure is on the order of $1 \times 10^{-5}$. The three fitting conditions are summarized in table 3, but these three conditions are simply three different values of our free parameter $K_E$ and its complement $K_C$. While some parameters of the tendon are non-physical, these parameters have no impact on the overall deformation. Thus, they are set for stable integration. All of the above simulations were run in $\approx$ 5 minutes on a standard desktop computer with 8Gb of RAM. The speed of these simulations can be attributed to the stability of the implicit integrator at large timesteps as well as the optimizations addressed throughout the process such as defining the jacobian pattern of the implicit system as well as choosing appropriate damping constants.

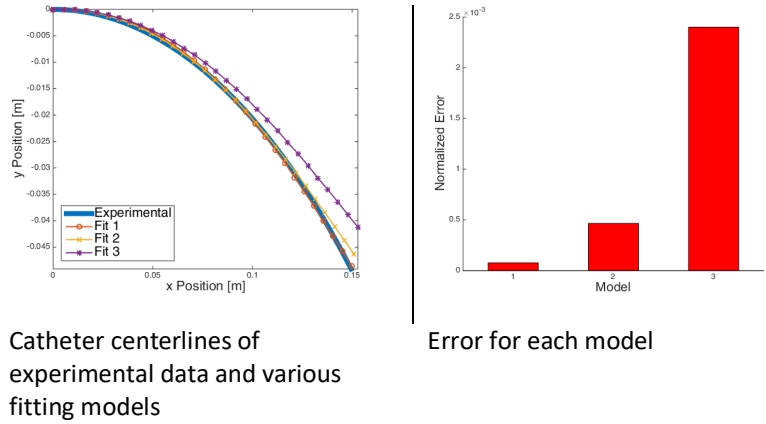

Catheter centerlines of experimental data and various fitting models

Error for each model

**Figure 22:** Experimental Data with Various Cosserat Models

To generalize the tendon configuration, the construction of the constraint given by equation (26) could to be extended as a combination of both vectors $\vec{d_1}$ and $\vec{d_2}$ as expressed by Eq. (32).

$$\vec{L_i} = r_L \vec{d_{t_i}} + \vec{r_i} \quad (32)$$

$\vec{d_t} = \alpha \vec{d_1} + \beta \vec{d_2}$ such as $\parallel \vec{d_t} \parallel = 1$ with coefficients $\alpha$ and $\beta$ weighting the influence of the basis vectors. It is a combination of both axial axes of the catheter to ensure that any point out of the lumen is brought back to the lumen line. However, we are indeed making an assumption that the points remain close enough to the plane given by $\vec{d_t d_3}$. In the general case, the constraint path is supported by $\vec{d_t}$ instead of $d_1$ (See Fig. 23). Then equations (25),(26),(27) become respectively: (33), (34) and (35).

$$(\vec{F_L})_i = [K_L (C_L)_j](\vec{d_t})_j \quad (33)$$
$$(C_L)_j = (\vec{L_j} - \vec{p_i}) \cdot (\vec{d_t})_j \quad (34)$$
$$(\vec{F_L})_i = [K_L (\vec{L_j} - \vec{p_i}) \cdot (\vec{d_t})_j](\vec{d_t})_j \quad (35)$$

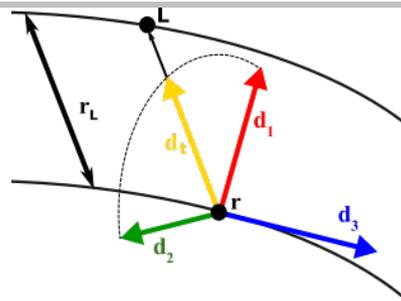

**Figure 23:** Tendon constraint path in the general case

Conclusions

In this work, we have implemented and validated a robotic catheter modeling framework using Cosserat rods as the 1D models for both the actuation tendon and the catheter body. We first demonstrated the physical accuracy of a single Cosserat rod modeling a single catheter by validating against both an analytical model and experimental data. Next we expanded the single catheter system by introducing a second Cosserat rod as the tendon and implementing a series of penalty force constraints to tie the two rods together. We then validated tendon-catheter system against experimental data of a real catheter, demonstrating its physical accuracy.

Moving forward, this model should be further validated against other experimental conditions to demonstrate its generalizability. Once fully validated, we can then take advantage of the mechanical nature of the model to explore various simulations that would not be possible with a geometrically-based model. Another interesting application is simulating catheter tip collision with tissue. Given that we know the forces on the control points at all times simply from the catheter position, it should be possible to determine the force the catheter is applying on tissue simply from its 3D position [15, 16]. In a cardiac procedure scenario, we may have access to the 3D position of the rod from imaging, but be unsure of the forces on the endpoint. A mechanical model such as ours would allow predictions of the endpoint force, thereby allowing the controller to prevent tissue injury or maintain sufficient pressure during ablations. Lastly, we can further extend the model to include $n$ tendons to explore more degrees of freedom. This will be a simple expansion of our current framework, given that all the new constraints would simply be computed as penalty forces and applied to our net force vector. All of these extensions are made possible by the mechanical nature of our model. The expansion of our framework will be done in C++ to reach real-time computation, which is not the case in the current prototype implemented in MATLAB.

Disclosures

The authors have no relevant financial interests in the manuscript and no other potential conflicts of interest to disclose.